\documentclass{article}

\PassOptionsToPackage{numbers, compress}{natbib}
 \usepackage[preprint]{neurips_2025}

\usepackage[utf8]{inputenc} 
\usepackage[T1]{fontenc}    
\usepackage{hyperref}       
\usepackage{url}            
\usepackage{booktabs}       
\usepackage{amsfonts}       
\usepackage{nicefrac}       
\usepackage{microtype}      
\usepackage{xcolor}         

\usepackage{graphicx}

\newcommand{\mathmat}[1]{\mathbf{#1}}

\newcommand{\mathfuc}[1]{\mathtt{#1}}

\newcommand{\smallplus}{{\scriptstyle +}}
\newcommand{\smallminus}{{\scriptstyle -}}
\newcommand{\tinyplus}{{\scriptscriptstyle  +}}
\newcommand{\tinyminus}{{\scriptscriptstyle  -}}

\newcommand{\myPara}[1]{\vspace{0pt}\noindent\textbf{#1}}

\usepackage{multirow}
\usepackage{bbding}
\usepackage{pifont}
\usepackage{wasysym}
\usepackage{comment}

\usepackage{makecell}

\newcommand{\sbd}{$\clubsuit$~}
\newcommand{\cl}{$\spadesuit$~}

\newcommand{\sbdf}{\textcolor{blue}{$\clubsuit$}~}
\newcommand{\clf}{\textcolor{blue}{$\spadesuit$}~}

\newcommand{\figref}[1]{Fig.~\ref{#1}}
\newcommand{\tabref}[1]{Tab.~\ref{#1}}
\newcommand{\secref}[1]{Sec.~\ref{#1}}

\usepackage{booktabs}
\usepackage{enumitem}

\title{RefCut: Interactive Segmentation with Reference Guidance}

\author{%
Zheng Lin$^1$ \quad Nan Zhou$^1$\thanks{Internship at Tsinghua University.} \quad Chen-Xi Du$^{1}$ \quad Deng-Ping Fan$^2$ \quad Shi-Min Hu$^1$\\
$^1$Tsinghua University \ \ \  $^2$Nankai University
}

\begin{document}

\maketitle
\vspace{-20pt}

\begin{abstract}
%
Interactive segmentation aims to segment the specified target on the image with positive and negative clicks from users.
Interactive ambiguity is a crucial issue in this field, which refers to the possibility of multiple compliant outcomes with the same clicks, such as selecting a part of an object versus the entire object, a single object versus a combination of multiple objects, and so on.
The existing methods cannot provide intuitive guidance to the model, which leads to unstable output results and makes it difficult to meet the large-scale and efficient annotation requirements for specific targets in some scenarios.
To bridge this gap, we introduce RefCut, a reference-based interactive segmentation framework designed to address part ambiguity and object ambiguity in segmenting specific targets. 
Users only need to provide a reference image and corresponding reference masks, and the model will be optimized based on them, which greatly reduces the interactive burden on users when annotating a large number of such targets.
In addition, to enrich these two kinds of ambiguous data, we propose a new Target Disassembly Dataset which contains two subsets of part disassembly and object disassembly for evaluation.
In the combination evaluation of multiple datasets, our RefCut achieved state-of-the-art performance.
Extensive experiments and visualized results demonstrate that RefCut advances the field of intuitive and controllable interactive segmentation. 
Our code will be publicly available and the \textcolor{black}{demo video} is in \href{https://www.lin-zheng.com/refcut}{https://www.lin-zheng.com/refcut}.
\end{abstract}
\vspace{-15pt}


\section{Introduction}
\label{sec:intro}
%
Interactive Segmentation (IS)~\cite{ramadan2020iis_survey} has gained significant attention with the advancement of deep learning techniques.
It allows users to guide the model in segmenting images to the target mask by providing simple inputs like clicks or scribbles, enabling more controllable, accurate, and precise results.
%
%
This task is especially valuable across various domains: 
in image annotation, it provides essential training labels~\cite{benenson2019human_annotators} for various segmentation tasks, such as semantic segmentation~\cite{chen2018deeplabv3p}, instance segmentation~\cite{he2017maskrcnn}, and salient object detection~\cite{borji2015sod_benckmark};
in artificial intelligence generated content, it supports precise object masks, facilitating creative processes for these mask-based methods~\cite{rombach2022high,zhang2023adding,chen2024anydoor};
and in medical imaging, it assists in precisely identifying target areas like organ or lesions~\cite{wang2018deepigeos,marinov2024deep}. 
\begin{figure}[t]
    \centering
    \includegraphics[width=1.0\linewidth]{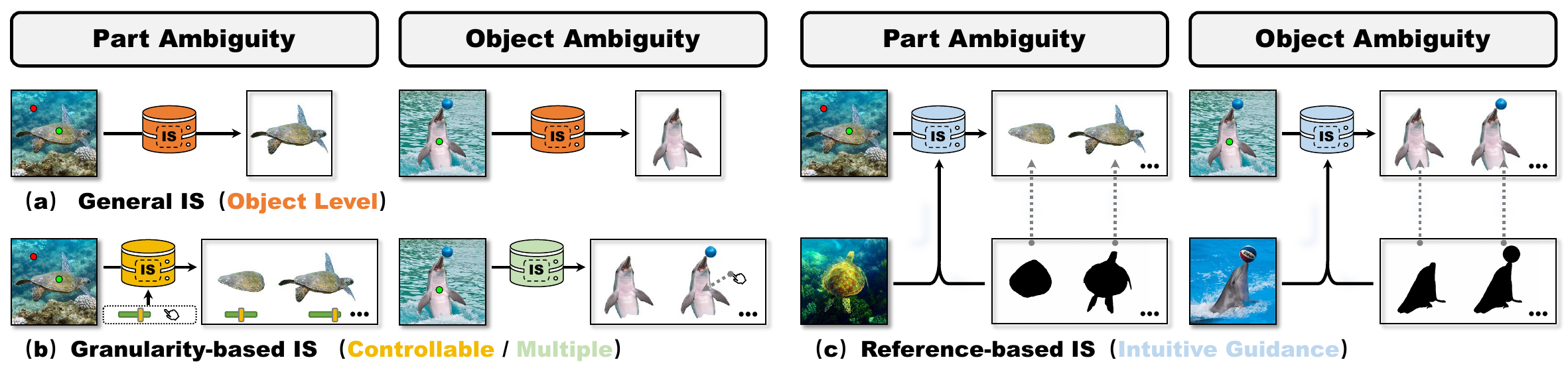}
    \caption{
    Comparison between our RefCut and other methods for Interactive Segmentation (IS). 
    (a) The general IS methods focus on the object level;
    (b) The granularity-based IS methods need selecting targets or inputting granularity values, which is universal but lacks certain stability;
    (c) Our reference-based IS method can provide more intuitive information to efficiently resolve part ambiguity and object ambiguity under specific requirements.}
    \label{fig:introduction}
    \vspace{-10pt}
\end{figure}

For interactive segmentation, various interaction modes have been explored, such as bounding boxes~\cite{lempitsky2009bounding_box_prior,wu2014milcut,xu2017deepgrabcut,zhang2020iog,zhang2022iog_pami}, polygons~\cite{li2004lazy_snapping,castrejon2017polygonrnn,acuna2018polygonrnnpp,ling2019curvegcn}, and scribbles~\cite{agustsson2019full_iis,liu2024segnext,bai2014error_tolerant_scribbles,han2022slim_scissors}. 
However, the click-based approaches~\cite{le2018boundary_prediction,maninis2018dextr,papadopoulos2017extreme_clicking,wang2019delse,liew2021thin,jain2019click_carving} are increasingly popular due to its ease of use, especially the positive and negative clicks~\cite{song2018seednet,hao2021edgeflow}, and this work adopts this interaction mode as well.
%
Recently deep learning-based techniques have become dominant.
Researchers are systematically investigating this task from multiple perspectives: 
utilizing the characteristics of different clicks~\cite{lin2020fcanet,lin2022focuscut}, designing network architectures~\cite{liu2023simpleclick,hu2019fctsfn}, optimizing the parameters online~\cite{kontogianni2020continuous_adaptation,lin2023siis}, focusing on more precise results~\cite{chen2022focalclick,liu2024segnext}, and so on.

Among above perspectives, one critical issue is the problem of interactive ambiguity. 
When a user provides the clicks on the target, multiple plausible targets may match the interactive points, creating challenges in accurately identifying the user’s desired one. 
As shown in \figref{fig:introduction}, the interactive ambiguity can be classified into two main types: part ambiguity and object ambiguity. 
The part ambiguity refers to the inability of the model to determine whether the user wants to segment a local part of the object or the entire object, such as the turtle’s shell versus the whole turtle.
The object ambiguity refers to the inability of the model to determine whether the user wants to segment one object or the combination of multiple objects, such as a dolphin alone versus a dolphin with a ball on its head.
For most general IS methods (\figref{fig:introduction}-a), they tend to solve the problem of whole object segmentation.
Some granularity-based methods (\figref{fig:introduction}-b) have also been proposed for the ambiguity problem, which fall into two main categories:
The granularity-controllable method~\cite{zhao2024graco} needs the user to input a value for the granularity, which requires several manual attempts.
The multi-granularity methods~\cite{li2018latent_diversity,liew2019multiseg} will provide multiple possible results that require the model to automatically select or the user to manually select the desired one.
These above methods are important for granularity segmentation in general-purpose situations. 
While these methods attempt to mitigate interactive ambiguity, they still may yield unstable results.
For scenarios where the user requires large-scale segmentations of specific part or some combinations of objects, such as the horse head or the rider with the horse, providing more intuitive and effective guidance can help to obtain more stable results.

To address the challenge of interactive ambiguity for specific targets, we introduce RefCut, an interactive segmentation framework based on the reference image.
In our approach, users provide an extra image of the same type as the targets along with its corresponding mask. 
The model utilizes a shared backbone to extract target-specific features from the reference, which serves as a prompt for the interactive segmentation network. 
This guidance enables the model to achieve a result that matches the reference in both granularity and semantics with minimal user input. 
In addition, users can also provide the negative mask to indicate regions or semantic targets that should be suppressed, enhancing the model’s ability to ignore unwanted targets. 
To evaluate the ambiguity-related performance of interactive segmentation methods, we also collect and annotate a new \textbf{T}arget \textbf{D}is\textbf{a}ssembly (TDA) dataset, which includes two subsets focusing on part-based and object-based ambiguities.
Our method demonstrates strong performance on PartImageNet~\cite{he2022partimagenet}, PASCAL-Part~\cite{chen2014pascalpart}, and the proposed TDA dataset.
%
%
It means that RefCut advance controllable interactive segmentation for specific targets with low ambiguity, which has significant importance in practical use.

The contributions can be summarized as follows:

\begin{itemize}[leftmargin=20pt,topsep=0pt, parsep=0pt]
	\item[$\bullet$] We propose RefCut, which to our knowledge is the first IS framework based on image reference to address part and object ambiguity in segmenting specific targets.
	\item[$\bullet$] We propose the TDA dataset that includes two subsets for part ambiguity and object ambiguity, to promote the development of this field for interactive ambiguity.
	\item[$\bullet$] Our method achieves SOTA on multiple datasets about ambiguity, indicating its ability to efficiently annotate a large number of specific objects in practical applications.
\end{itemize}



\vspace{-10pt}
\section{Related Work}
\vspace{-5pt}
\subsection{General Interactive Segmentation}
Traditional interactive segmentation methods are mainly based on the low-level features of the image and use traditional algorithms~\cite{adams1994seeded_region_growing,mortensen1995intelligent_scissors,kim2010nonparametric}, such as graph cut~\cite{boykov2006graphcut_ijcv,blake2004adaptive_gmmrf,rother2004grabcut,vicente2008graphcut_connectivity_priors,veksler2008star_graphcut}, random walks~\cite{grady2006random_walk,kim2008rwr_restart}, geodesic distance~\cite{gulshan2010geodesic_star,bai2009geodesic_matting,price2010geodesic_graphcut}, etc., to differentiate between pixels.
DOS~\cite{xu2016dos} presents the first deep learning-based method with specification, and subsequent researchers explored it in different aspects.
\textbf{(1) For improving training paradigms},
ITIS~\cite{mahadevan2018itis} and RITM~\cite{sofiiuk2022ritm} optimize the training process by introducing iterative training;
SimpleClick~\cite{liu2023simpleclick} proposes a simple yet effective framework based on ViT~\cite{dosovitskiy2020vit} architecture;
MIS~\cite{li2023mis} adopts an unsupervised method to generate rich data for training.
%
\textbf{(2) For better utilizing the interaction information},
CMG~\cite{majumder2019cmg} transforms the click information into a content-aware map;
FCANet~\cite{lin2020fcanet} and FocusCut~\cite{lin2022focuscut} distinguish between the localization and refinement roles of initial and subsequent clicks;
CDNet~\cite{chen2021cdnet} and GPCIS~\cite{zhou2023gpcis} utilize the non-local method and Gaussian process to propagate click information to the full map;
PseudoClick~\cite{liu2022pseudoclick} uses the network to simulate human-generated clicks;
CPlot~\cite{liu2024cplot} introduces click prompt learning with optimal transport to delve deeper into the information of each click.
 %
%
\textbf{(3) For better segmentation with precise details}, 
RIS-Net~\cite{liew2017risnet}, FocusCut~\cite{lin2022focuscut}, FocalClick~\cite{chen2022focalclick}, and FCFI~\cite{wei2023fcfi} repair details focusing on local views from click pairs or a single click;
CFR-ICL~\cite{sun2024cfr-icl} proposes cascade-forward refinement and  MFP~\cite{lee2024mfp} makes full use of probability maps for more precise results.
%
%
\textbf{(4) For optimizing the model},
BRS~\cite{jang2019brs} and f-BRS~\cite{sofiiuk2020fbrs} utilize the backpropagating refinement method for the input map and the network features;
IA-SA~\cite{kontogianni2020continuous_adaptation} and SIIS~\cite{lin2023siis} continuously optimize the model parameters using the information of the clicks and the generated mask during the interactive process, respectively.
\textbf{(5) For a lower latency},
Interformer~\cite{huang2023interformer}, EMC-Click~\cite{du2023emc_click}, FDRN~\cite{zeng2023fdrn} and SegNext~\cite{liu2024segnext} eliminate the need to repeatedly process time-consuming feature extraction processes.
%

%

\vspace{-5pt}
\subsection{Granularity-based Interactive Segmentation}
\vspace{-5pt}
Some related works have also explored the problem of interactive ambiguity. These granularity-based methods are mainly divided into two categories.
The multi-granularity methods tend to generate multiple possible outputs~\cite{guzman2012multiple_choice_learning,li2023semantic_sam}.
LD~\cite{li2018latent_diversity} uses a convolutional neural network to generate multiple possible outcomes and another one to automatically pick the final result from them;
MutiSeg~\cite{liew2019multiseg} is a scale-diverse interactive image segmentation model that incorporates scale priors to generate segmentation outputs of varying scales, allowing users to efficiently locate desired segmentations with reduced input.
The popular SAM~\cite{kirillov2023sam} can also be seen as a multi-granularity method, which generates results of various uncertain granularities for different targets in the image in its global mode. 
And its interactive mode is similar to general interactive segmentation methods.
The granularity-controllable methods tend to enable users to more effectively control granularity.
GraCo~\cite{liew2019multiseg} enables precise control over segmentation granularity through adjustable input values, enhancing customization and resolving spatial ambiguity.
However, these methods are mainly aimed at generalized interactive segmentation situations, and if the user needs to segment a specific type of target, more intuitive information could be introduced to give the model a better indication of what to expect. 
This type of approach has not been explored in depth, which is our main concern in this work.
%





\vspace{-10pt}
\section{The Proposed Method}
\vspace{-5pt}

\label{sec:method}

\subsection{Overall Framework}
\begin{figure*}[t]
    \centering
    \includegraphics[width=1.0\linewidth]{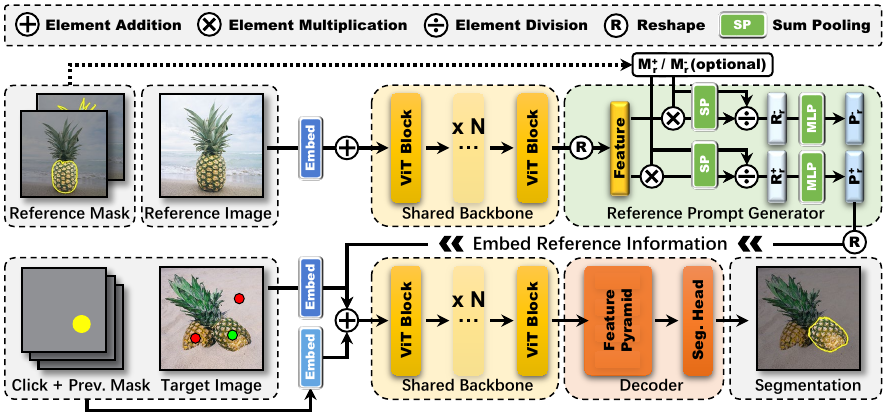}
    \caption{The framework of RefCut. The upper branch shows the extraction process of the reference prompts from the reference image and masks, and the lower branch shows the classic interactive segmentation pipeline from the target image and user clicks.}
    \label{fig:framework}
    \vspace{-10pt}
\end{figure*}

As shown in \figref{fig:framework}, our RefCut framework is built based on the SimpleClick~\cite{liu2023simpleclick} framework, which has been the basis for many works~\cite{lee2024mfp,zhao2024graco} in recent years.
The lower target branch shows the basic pipeline for interactive segmentation.
The upper reference branch shows the process for RefCut to get the reference prompts.
The following will introduce these two branches separately.

\myPara{Target Branch.}
The target image $\mathmat{I}_{t}$ is fed into a patch embedding module $\mathfuc{E}_{img}$ for the image.
The positive and negative clicks are transformed into two disk maps.
Then the maps are concatenated with the previous prediction mask to constitute the extra maps $\mathmat{C}_{t}$, which is fed into an extra patch embedding module $\mathfuc{E}_{ext}$.
For the baseline, the embedded feature $\mathmat{F}_e$ is the sum of the above two:
\begin{equation} 
\mathmat{F}_e=\mathfuc{E}_{img}\left(\mathmat{I}_{t}\right) + \mathfuc{E}_{ext}\left(\mathmat{C}_{t}\right).
\end{equation}
This feature is input into the backbone and decoder in order to obtain the final segmentation $\mathmat{S}_t$:
\begin{equation} 
\mathmat{S}_t=\mathfuc{Decoder}\left(\mathfuc{Backbone}\left( \mathmat{F}_e\right)\right).
\end{equation}
The backbone is the ViT~\cite{dosovitskiy2020vit} and the decoder contains a simple feature pyramid~\cite{li2022exploring} and  a segmentation head, which is detailed in ~\cite{liu2023simpleclick}.
For the training phase, the segmentation result will calculate the loss with the ground truth to optimize the model parameters.

\myPara{Reference Branch.}
The users need to provide a reference image and corresponding guidance masks $\mathmat{M}^{*}_{r}$, which are one or two 0-1 maps.
The positive mask $\mathmat{M}^{\tinyplus}_{r}$ indicates the foreground pixels of the reference target, such as the object part or the object combination.
The negative mask $\mathmat{M}^{\tinyminus}_{r}$ indicates the  pixels that need to be suppressed, such as the  remaining parts or the object that needs to be ignored.
The negative mask is optional.
If users want convenience, they can set it to empty.
The reference image $\mathmat{I}_{r}$ is fed into the same patch embedding module $\mathfuc{E}_{img}$.
Then the output will be solely input into the same backbone as the target branch to obtain the reference image feature $\mathmat{F}_{r}$:
\begin{equation} 
\mathmat{F}_r=\mathfuc{Backbone}\left(\mathfuc{E}_{img}\left(\mathmat{I}_{r}\right)  \right)
\end{equation}
With the guidance masks $\mathmat{M}^{*}_{r}$ and $\mathmat{F}_{r}$, the Reference Prompt Generator ($\mathfuc{RPG}$) will get the reference prompts $\mathmat{P}^{*}_{r}$:
\begin{equation} 
\mathmat{P}^{*}_{r}=\mathfuc{RPG}\left(\mathmat{M}^{*}_{r},\mathmat{F}_{r}\right), {*}\in\{\smallplus,\smallminus\}.
\end{equation}
For RefCut, the reference prompts will be add to the embedded feature $\mathmat{F}_e$, which are fed into subsequent modules to get the final segmentation:
\begin{equation} 
\mathmat{F}_e=\mathfuc{E}_{img}\left(\mathmat{I}_{t}\right) + \mathfuc{E}_{ext}\left(\mathmat{C}_{t}\right) + \mathmat{P}^{\tinyplus}_{r} + \mathmat{P}^{\tinyminus}_{r}.
\end{equation}

\subsection{Reference Prompt Generator}
The upper right part of \figref{fig:framework} shows the calculation process of the reference prompts.
This process is divided into representation extraction and 
dual prompt encoding.

\myPara{Representation Extraction.}
In order to obtain positive and negative representations of the reference target, we need to selectively extract image features based on the reference mask $\mathmat{M}^{*}_{r}$.
We first reshape the feature extracted from the ViT backbone into the standard convolutional feature shape ($H'\times W'\times C_0$ ) to get the reference image feature $\mathmat{F}_{r}$.
The $H'$ and $W'$ of this feature are $\frac{1}{16}$ of the original $H$ and $W$ of the reference image $\mathmat{I}_{r}$.
Here, the reference masks $\mathmat{M}^{*}_{r}$ have also been resized to the same size.
We first multiply the reference image feature with the reference mask and perform sum pooling on them to form a one-dimensional feature.
Afterwards, this feature will be divided by the value from the sum pooling on the reference mask to obtain the final reference target representation $\mathmat{R}^{*}_{r}$.
It is formulated as:
\begin{equation} 
\mathmat{R}^{*}_{r}=  \frac{\sum \left(\mathmat{F}_{r}\otimes\mathmat{M}^{*}_{r}\right)}{\sum \mathmat{M}^{*}_{r}}, {*}\in\{\smallplus,\smallminus\},
\end{equation}
where $\otimes$ means the element multiplication, $\sum$ means the sum pooling operation.
According to whether the reference image mask is positive or negative, the obtained representations are divided into $\mathmat{R}^{\tinyplus}_{r}$ and $\mathmat{R}^{\tinyminus}_{r}$.

\myPara{Dual Prompt Encoding.}
The previously obtained representations only correspond to the representations of the reference target within the mask.
In order to activate the positive representation and suppress the negative representation in the model. 
We encode these two representations using two layers of Multi-Layer Perceptron ($\mathfuc{MLP}$) modules with the same structure but different parameters to obtain the final prompts $\mathmat{P}^{*}_{r}$:
\begin{equation} 
\mathmat{P}^{*}_{r}=\mathfuc{MLP}\left( \mathmat{R}^{*}_{r}\right), {*}\in\{\smallplus,\smallminus\}.
\end{equation}
Finally, these two prompts will be added to the embedded feature of the baseline as the reference.

\subsection{Pair Data Sampling}
\label{sec:sampling}
To train the model of RefCut, we need a large number of target and reference image pairs.
The targets on these pairs should have common semantics.
We select the PartImageNet~\cite{he2022partimagenet} dataset which includes 158 categories and part segmentation for training and evaluation.

\myPara{Sampling in Training.}
For one sample, We first randomly select an object with a target image.
Assuming this object is divided into $N$ kinds of parts, we randomly select $1\sim N$ parts from them.
The masks of these parts will be merged together to obtain the ground truth of the target.
Afterwards, we will select an object with the same category from the dataset as the reference image.
The same parts on the reference object are merged together to obtain the positive reference mask $\mathmat{M}^{\tinyplus}_{r}$.
The remaining unselected parts are merged to get the negative reference mask $\mathmat{M}^{\tinyminus}_{r}$.

\myPara{Sampling in Evaluation.}
In order to make the evaluation more reflective of the performance of the reference guidance, rather than due to the network fitting to the segmentation of parts, our evaluation adopts a combination evaluation.
For an object, we will evaluate its segmentation performance for a single part, the combination of multiple parts, and the whole object that includes all parts.
Due to the diversity of combinations, our selection criteria for the combination evaluation is to choose all combinations of two parts whose masks must be connected.
For the stable and unique selection of the reference sample during evaluation, we first select all the samples in the evaluation set that can be used as references, and then choose the next sample with a file name in lexicographic order compared to the current sample as our reference sample.
Afterwards, the same method as training is used to obtain the input of the network.

\subsection{Target Disassembly Dataset}

\label{sec:dataset}
In order to effectively evaluate the reference-based method, we need a dataset that can reflect the performance of both part ambiguity and object ambiguity. 
For part ambiguity, existing datasets such as PartImageNet~\cite{he2022partimagenet} and PASCAL-Part~\cite{chen2014pascalpart} can be used for evaluation to a certain extent, but both of them mainly contain categories such as animals and transportation, and lack some common disassemblable objects in life. 
For object ambiguity, there is no dataset specialized for object combinations, especially common connected object combinations. 
To fill these gaps, we propose the \textbf{T}arget \textbf{D}is\textbf{a}ssembly (TDA) Dataset.
It took one month for us to carefully select representative data which is different from that in other datasets.
We get relevant objects or combinations through observation, thinking, consulting AI, and collect pictures with appropriate copyright that can be used for academic research, and label them manually.
In the end, there are a total of 400 images and disassembly annotations, which can be divided into two distinct subsets: the \textbf{P}art \textbf{D}isassembly set (TDA-PD) and the \textbf{O}bject \textbf{D}isassembly set (TDA-OD). 

%

\begin{figure*}[t]
    \centering
    \includegraphics[width=1.0\linewidth]{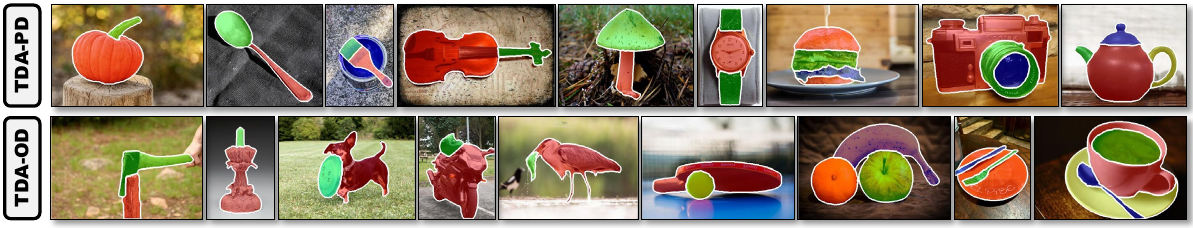}
    \caption{The examples of Target Disassembly Dataset with subsets for part disassembly (TDA-PD) and object disassembly (TDA-OD). }
    \label{fig:dataset}
    \vspace{-10pt}
\end{figure*}

\myPara{Part Disassembly Set.}
The TDA-PD subset focuses on resolving part-level ambiguity within single objects. 
It includes 50 unique objects, each with 5 annotated images. 
For each object, we provide both images and detailed segmentation masks that label individual parts with distinct tags based on their function or structure, such as the leaves and body of a pineapple or the head and handle of a hammer. 
Each part type is consistently tagged across instances, facilitating accurate selection for reference. 
This subset mainly contains common targets in our daily life, and can also be used as an extension to other part datasets to verify category generalization for these part-related tasks.

\myPara{Object Disassembly Set.}  
The TDA-OD subset addresses object-level ambiguity in complex object combinations.  
It includes 30 unique object combinations, each with 5 annotated images.  
In this subset, we provide images and detailed segmentation masks for each combination, labeling each distinct object within a scene with consistent tags, such as the dog and the frisbee it bites, the pen and the notebook it writes in, and so on.
Each object type within these combinations is also consistently tagged.
%
%
The TDA-OD subset is designed to capture intricate multi-object relationships commonly seen in daily life but rarely represented in existing datasets, making it a valuable resource for tasks that require understanding complex object interactions.

\vspace{-10pt}
\section{Experiment}
\label{sec:exp}
\vspace{-5pt}

\subsection{Experiment Settings}
\label{sec:setting}

\myPara{Datasets.} The commonly used IS datasets~\cite{rother2004grabcut,mcguinness2010berkeley,hariharan2011sbd,perazzi2016davis} focus on the segmentation of individual objects.
They cannot provide the situations of object parts and combinations, and some of them are unable to provide semantic information to search for reference images.
Therefore, we selected the following datasets with CC-BY 4.0 license referring to \cite{zhao2024graco} for our experiments:
\begin{itemize}[leftmargin=20pt,topsep=0pt, parsep=0pt]
	\item[$\circ$] \texttt{PartImageNet}~\cite{he2022partimagenet}: It contains 24080 images from ImageNet~\cite{deng2009imagenet}, with training, validation, and testing of 20466, 1206, and 2408 images, respectively. It involves 158 categories disassembled into 2$\sim$5 parts. Due to its category diversity, we use its training set for training and the testing set for evaluation.  
    The number of samples for combination evaluation is 14495.

	\item[$\circ$] \texttt{PASCAL-Part}~\cite{chen2014pascalpart}:
    It is an extension of the PASCAL VOC~\cite{everingham2010pascal} dataset regarding parts.
    It involves only 20 categories and finer granularity, making it more suitable for evaluating granularity-based methods~\cite{zhao2024graco} that are independent of categories. 
    The difference set between the validation set and the SBD~\cite{hariharan2011sbd} training set is used for evaluation.  
    All objects with part annotation are evaluated.
    The number of samples for combination evaluation is 106699.

	\item[$\circ$] \texttt{TDA (Ours)}:
    It contains 400 images with 80 categories covering both part ambiguity and object ambiguity. 
    The number of samples for combination evaluation is 1336.
    %
    
\end{itemize}

\myPara{Evaluation Metrics.}
We perform the evaluation following the standard protocol used in previous click-based IS methods.
Specifically, the initial positive click is placed at the center of the target, while subsequent clicks are positioned in the largest error regions identified by comparing the current prediction to the ground truth.
For evaluation metrics, we use the Number of Clicks (NoC) to measure performance, which calculates the average number of clicks needed to reach a fixed Intersection over Union (IoU) threshold, with lower values indicating better performance.
The maximum number of clicks is set to 20.
We provide the results with IoU thresholds of 80\%, 85\%, and 90\%, which is shown as NoC@X.
Due to the difference in segmentation between parts and objects, there is significant uncertainty, especially for animal parts. 
Therefore, the thresholds of 80\% and 85\% are more informative for the parts.
In ablation study, the IoU for the initial click (IoU\&1) is also provided to verify the effect of introducing reference guidance.
Unlike typical IS, as discussed in \secref{sec:sampling}, we adopt a combination evaluation.
For example, we may evaluate the head, head with body, body with tail and the whole object for fish with 3 parts, which reflect the performance about ambiguity.

\myPara{Implementation Details.}
We build RefCut based on the SimpleClick~\cite{liu2023simpleclick} framework on 1 NVIDIA RTX 3090Ti. 
In the experiments, we mainly use the ViT-B~\cite{dosovitskiy2020vit} model trained on the SBD~\cite{hariharan2011sbd} and COCO~\cite{lin2014mscoco}+LVIS~\cite{gupta2019lvis} datasets as our base model, and finetune it using the PartImageNet~\cite{he2022partimagenet} dataset with rich categories.
Like \cite{liu2023simpleclick}, we adopt data augmentation strategies such as random scaling, cropping, flipping, etc., and use normalized focal loss~\cite{sofiiuk2022ritm} to train our network.
We train RefCut for 55 epochs using the Adam~\cite{kingma2014adam} optimizer with $\beta_1 = 0.9$ and $\beta_2 = 0.999$, starting with a learning rate of $5e^{-6}$ that decayed by a factor of 10 after 50 epochs.
In order to increase the richness of reference masks, we have a 25\% probability of only positive or negative reference mask.
%
%

\subsection{Result \& Analysis}
\begin{figure*}[t]
    \centering
    \includegraphics[width=1.0\linewidth]{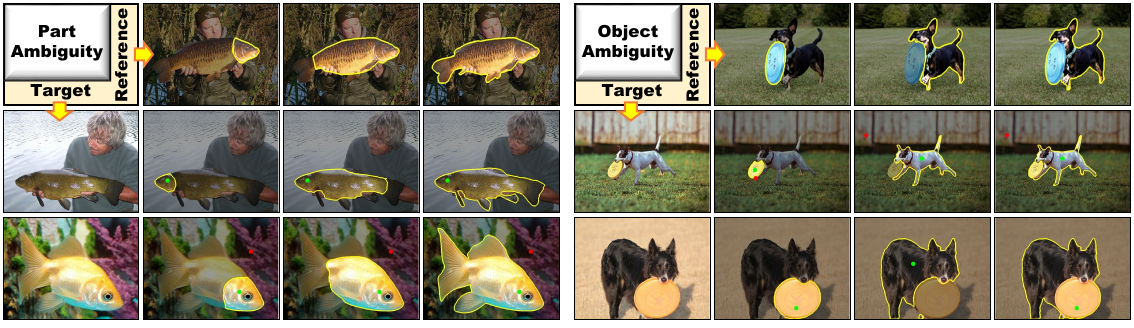}
    \vspace{-20pt}
    \caption{The visualized results. 
    %
    %
    The yellow contours in the reference and target images represent the reference masks and segmentation results, respectively.
    %
    %
    The \textbf{baseline results} are similar to those obtained using a single object as a reference (4th, 6th, 7th cloumns). 
    }
    \label{fig:result}
    \vspace{-5pt}
\end{figure*}

In this section, \tabref{tab:sota} and \figref{fig:result} show the quantitative and visualized results of RefCut.
\tabref{tab:oss-is} compares RefCut with One-Shot Segmentation (OSS) and provides an analysis.


\begin{table*}[t]
	\centering
	\renewcommand{\arraystretch}{0.95}
    \renewcommand{\tabcolsep}{3.2pt}
    \newcommand{\pasbd}{\scalebox{1.2}{\ding{169}}~}

    \newcommand{\reim}{$^{\dag}$}
    \newcommand{\noc}[1]{\small{NoC@#1}}
    \renewcommand{\noc}[1]{\small{@#1}}
    \newcommand{\method}[5]{{#1}{#2}~{#3}&{#5}}
	\begin{tabular}{l|c|ccc|ccc|ccc}
	\toprule
	\multirow{2}{*}{Method}  &\multirow{2}{*}{Backbone}   & \multicolumn{3}{c|}{PartImageNet} & \multicolumn{3}{c|}{PASCAL-Part} & \multicolumn{3}{c}{TDA}  \\ \cline{3-11} 
	                       &                           &\noc{80} &\noc{85} &\noc{90} &\noc{80} &\noc{85} &\noc{90} &\noc{80} &\noc{85} &\noc{90}      \\ 
    \midrule
    \method{\pasbd}{SAM}{\cite{kirillov2023sam}}{ICCV23}{ViT-B}           &8.87 &11.25 &14.57 &12.60 &14.86 &17.31 &2.96 &4.16 &7.01 \\
    \midrule
    \method{\sbd}{CDNet}{\cite{chen2021cdnet}}{ICCV21}{ResNet-34}   &7.62 &9.49 &12.24 &11.47 &13.43 &15.92 &4.22 &5.29 &7.49 \\
    \method{\sbd}{RITM}{\cite{sofiiuk2022ritm}}{ICIP22}{HRNet-18}       &5.69 &7.28 &10.00 &8.32 &10.26 &13.30 &3.31 &4.02 &5.58 \\
    \method{\sbd}{FocusCut}{\cite{lin2022focuscut}}{CVPR22}{ResNet-101}  &5.25 &6.84 &9.66 &7.38 &9.35 &12.40 &2.86 &3.51 &5.16 \\
    \method{\sbd}{GPCIS}{\cite{zhou2023gpcis}}{CVPR23}{ResNet-50}     &5.20 &6.58 &9.08 &8.40 &10.26 &13.07 &2.84 &3.41 &4.84 \\
    \method{\sbd}{FCFI}{\cite{wei2023fcfi}}{CVPR23}{ResNet-101}     &5.89 &7.74 &10.78 &8.31 &10.45 &13.66 &3.51 &4.54 &6.56 \\
    \method{\sbdf}{SimpleClick}{\cite{liu2023simpleclick}}{ICCV23}{ViT-B}   &3.15 &4.08 &6.21 &6.94 &8.76 &11.76 &2.07 &2.47 &3.60 \\ 
    \method{\sbdf}{GraCo}{\cite{zhao2024graco}}{CVPR24}{ViT-B}         &3.84 &5.15 &7.69 &6.52 &8.49 &11.73 &2.56 &3.04 &4.31 \\ \hline
    \sbdf RefCut (Ours) &ViT-B &\textbf{2.41} &\textbf{3.31} &\textbf{5.42} &\textbf{6.40} &\textbf{8.26} &\textbf{11.36} &\textbf{1.84} &\textbf{2.20} &\textbf{3.33} \\ 
   
    \midrule
    \method{\cl}{CDNet}{\cite{chen2021cdnet}}{ICCV21}{ResNet-34}          &5.33 &7.04 &10.04 &8.75 &10.97 &14.14 &2.25 &2.86 &4.53 \\
    \method{\cl}{RITM}{\cite{sofiiuk2022ritm}}{ICIP22}{HRNet-32}            &4.49 &5.79 &8.29 &7.22 &9.02 &12.03 &2.01 &2.41 &3.44 \\
    \method{\cl}{FocalClick}{\cite{chen2022focalclick}}{CVPR22}{SegF-B3}       &4.28 &5.45 &7.76 &7.73 &9.47 &12.24 &2.01 &2.39 &3.44 \\
    \method{\cl}{EMC-Click}{\cite{du2023emc_click}}{CVPR23}{HRNet-32}       &4.36 &5.63 &8.14 &7.58 &9.21 &12.09 &1.91 &2.28 &3.19 \\
    \method{\cl}{FCFI}{\cite{wei2023fcfi}}{CVPR23}{HRNet-18}          &4.67 &6.09 &8.71 &7.41 &9.28 &12.38 &2.14 &2.50 &3.63 \\
    \method{\cl}{SegNext}{\cite{liu2024segnext}}{CVPR24}{ViT-B}            &4.82 &6.13 &8.64 &8.16 &10.08 &13.10 &2.00 &2.38 &3.37 \\
    \method{\cl}{MFP}{\cite{lee2024mfp}}{CVPR24}{ViT-B}                &3.98 &5.12 &7.45 &6.90 &8.66 &11.61 &1.84 &2.13 &3.10 \\
    \method{\clf}{SimpleClick}{\cite{liu2023simpleclick}}{ICCV23}{ViT-B}        &3.06 &3.97 &6.02 &6.83 &8.59 &11.53 &1.95 &2.30 &3.37 \\ 
    \method{\clf}{GraCo}{\cite{zhao2024graco}}{CVPR24}{ViT-B}              &3.45 &4.60 &6.98 &\textbf{6.29}  &8.18 &11.31 &1.99 &2.40 &3.44 \\
    \hline
    \clf RefCut (Ours) &ViT-B &\textbf{2.35} &\textbf{3.20} &\textbf{5.26} &6.36&\textbf{8.14} &\textbf{11.14} &\textbf{1.72} &\textbf{2.03} &\textbf{3.06} \\ 
    \bottomrule
	\end{tabular}
	\caption{
      Quantitative comparison about NoC metric using combination evaluation.
      %
      %
      \sbd and \cl indicate that their models or the base models they used for finetuning are trained on SBD~\cite{hariharan2011sbd} and COCO~\cite{lin2014mscoco}+LVIS~\cite{gupta2019lvis} datasets, respectively.
      \pasbd shows the original SAM with clicks.
      The \textcolor{blue}{blue} color indicates using the PartImageNet~\cite{he2022partimagenet} dataset for finetuning.
      The \textbf{bold} value represents the best.
      }

	\label{tab:sota}
    \vspace{-12pt}
\end{table*}

\myPara{Quantitative Results.}
In contrast to traditional datasets used for interactive segmentation, such as GrabCut~\cite{rother2004grabcut}, Berkeley~\cite{mcguinness2010berkeley}, SBD~\cite{hariharan2011sbd}, and DAVIS~\cite{perazzi2016davis}, which lack disassemblable annotation information, we evaluate RefCut on the PartImageNet~\cite{he2022partimagenet}, PASCAL-Part~\cite{chen2014pascalpart}, and our proposed TDA datasets.
We benchmark our method against recent interactive segmentation methods with available code and suitable pretrained weights, creating a comprehensive benchmark, as shown in \tabref{tab:sota}. 
Given that our method uniquely incorporates reference images, making a fully fair comparison challenging, this benchmark serves to provide a relative performance evaluation. 
%
%
We further finetune the recent granularity-based method GraCo~\cite{zhao2024graco} and the baseline SimpleClick~\cite{liu2023simpleclick} with the same training data.
We evaluate the performance with single-part, dual-part combinations, and the whole target.
From these results in the table, our method achieves leading performance across all four datasets in most cases. 
For PartImageNet dataset, our method provides clear reference guidance and demonstrates substantial performance gains.
For PASCAL-Part dataset, our improvement is modest because the dataset contains a large number of small parts and favors methods with controllable granularity, which will also be discussed in \secref{sec:ablation}.
%
For TDA dataset, our performance is also leading, but in this dataset, both subsets evaluate the segmentation of many entire objects, which is friendly to general interactive segmentation and therefore their performance is also good.

\myPara{Visualized Results.}
In \figref{fig:result}, we show the visualized results generated by RefCut using different references as guidance.
The left column is for the part ambiguity. 
We choose the category of fish as an example and select a reference image and different reference masks, including the head, head with body, and the whole fish. 
To reflect the role of the reference, we set the clicks in the same position for the same image.
As can be seen, when we segment these images, even if all positive clicks are only located on the fish head, our model can still segment the same target as the reference mask to obtain the accurate result.
To confirm the class generalization of our method, we select fish species in the third row that are completely different from the reference image. 
It can be seen that the reference still plays a good guiding role.
The right column is for the object ambiguity. 
We choose a combination of dog and frisbee as an example and select individual objects and a mixture of two objects as reference.
As can be seen, if a single object is selected as a reference, our method is similar to general interactive segmentation, which can effectively segment individual objects. 
However, if we choose their common mask as a reference, no matter which object we click on, our model can treat the two as a whole and quickly segment them together.
The color of the frisbee in the target images we selected is completely different from that in the reference image, and the category of frisbee does not appear in the training set of PartImageNet~\cite{he2022partimagenet}. 
However, our method is still effective, indicating that RefCut has learned more ability about generalization.

\myPara{Comparison \& Analysis with OSS.}
While our version of IS shares similarities with One-Shot Segmentation (OSS), they are fundamentally two distinct tasks. 
Our contribution lies in introducing references to address ambiguity in IS, which has not be explored.
IS focuses on manual segmentation for accurate results and heavily relies on interactions. The reference serves only as an aid to resolve ambiguities. In training,  clicks are always present, whereas references might occasionally be absent with a small probability. Therefore, even without a reference, our method can still work, but without clicks, the output will inevitably be empty. Moreover, IS produces the mask for target instance rather than all targets in the entire image. Subsequent clicks are used to refine ambiguities and target details.
OSS focuses on automatic segmentation and heavily depends on the reference. Its results involve segmenting all targets in the entire image that are similar to the reference.
In \tabref{tab:oss-is}, we include the results from some OSS methods with the baseline IS method for further refinement. The leading performance of RefCut is also in line with expectations, because OSS segments all similar targets, and using clicks to refine the results can be labor-intensive, such as when removing unwanted instances and refining bad result.

\begin{table}[tbp]
\centering
\renewcommand{\arraystretch}{0.95}
\renewcommand{\tabcolsep}{15.9pt}
\begin{tabular}{c|cccc}
\toprule
   Method  & NoC@80 & NoC@85  & NoC@90 & IoU\&1    \\ \midrule
 PerSAM~\cite{zhang2024persam}+IS & 3.50  & 4.43  & 6.48  & 52.48 \\ 

 Matcher~\cite{liu2024matcher}+IS    & 2.89  & 3.84  & 5.93  & 64.99 \\ 

SLiMe~\cite{khani2024slime}+IS    & 2.77  & 3.75  & 5.90  & 64.23 \\ \hline

\sbdf RefCut (Ours)    & \textbf{2.41}  & \textbf{3.31}  & \textbf{5.42}  & \textbf{74.74} \\
\bottomrule
\end{tabular}
\caption{Comparison with OSS methods in PartImageNet~\cite{he2022partimagenet}. }
\label{tab:oss-is}
\vspace{-15pt}
\end{table}

\vspace{-10pt}
\subsection{Ablation Study}
\label{sec:ablation}
\begin{figure}[t]
    \centering
	\begin{minipage}[b]{0.4\linewidth}
		\includegraphics[width=1\linewidth]{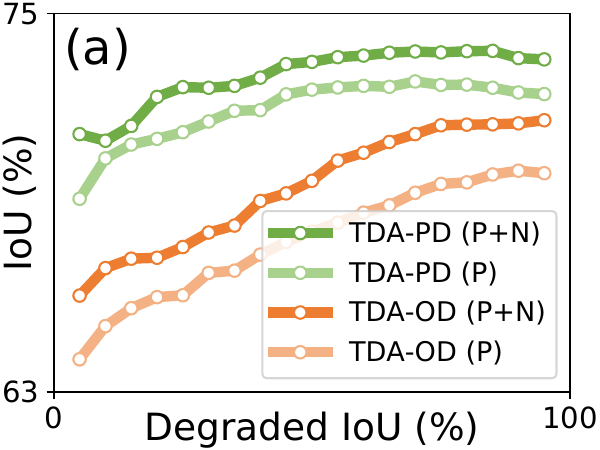}
	\end{minipage}
	\begin{minipage}[b]{0.4\linewidth}
		\includegraphics[width=1\linewidth]{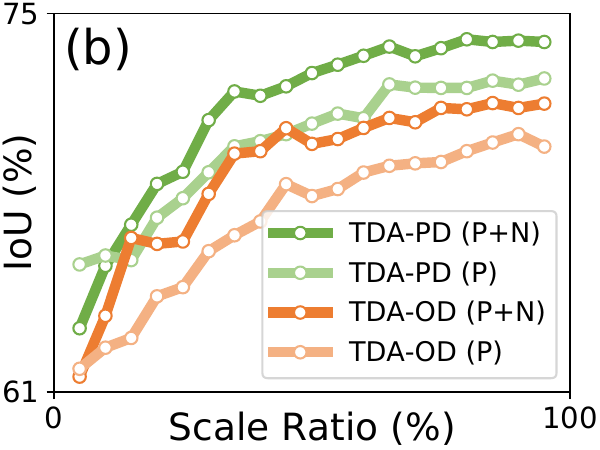}
	\end{minipage}
    \caption{The ablation study for the reference mask quality and target size on TDA subsets with RefCut based on SBD~\cite{hariharan2011sbd}. P and N mean adopting the positive and negative reference, respectively. 
    %
    }
    \label{fig:ablation}
    \vspace{-10pt}
\end{figure}

We conduct ablation study with various settings.
\tabref{tab:ablation} shows the situations for various reference.
%
%
\figref{fig:ablation} shows the influence about the quality of the reference mask and the size of the reference target.


\myPara{Reference Guidance.}
Providing the positive reference mask based on the reference image is the most intuitive way for users.
%
Compared to the baseline,
the positive guidance brings significant improvement.
In PartImageNet, the improvement for NoC exceeds 0.5 clicks, which is large for IS.
The IoU metric improvement of the first click more intuitively reflects the performance of the positive guidance.
On the three datasets, the performance improvements for the SBD-based version it brings are 13.85\%, 6.47\%, 5.81\%, respectively.
All of these clearly indicate that the positive reference mask can greatly reduce ambiguity and improve the initial segmentation results.
The negative reference mask is originally proposed to assist the positive one, but we also provide the performance of only the negative reference mask here. 
The improvement brought by the negative reference mask is not inferior to that of the positive reference mask at all.
%
%
This is also easy to understand, because the negative masks can suppress the targets that users do not want to segment, and the generation of interactive ambiguity is often due to excessive segmentation of certain parts.
We input both masks into RefCut and find that the performance is further improved.
But the value of this improvement is very limited.
This also indicates that some of the functions of positive and negative reference masks actually have a common impact.
Therefore, if the number of images that need to be annotated is not too large, users can choose only the positive reference mask.
%

\begin{table*}[htbp]
	\centering
	\renewcommand{\arraystretch}{0.95}
	\renewcommand{\tabcolsep}{2.9pt}

    \newcommand{\noc}[1]{\small{NoC@#1}}
    \renewcommand{\noc}[1]{\small{@#1}}
    \newcommand{\iou}[1]{\small{IoU\&#1}}
    \renewcommand{\iou}[1]{\small{\&#1}}
    \newcommand{\method}[5]{{#1}{#2}~\cite{#3}&{#5}}

	\begin{tabular}{c|c|c|cccc|cccc|cccc}
	\toprule
	\multirow{2}{*}{\#}  &Pos.  &Neg. & \multicolumn{4}{c|}{PartImageNet} & \multicolumn{4}{c|}{PASCAL-Part} & \multicolumn{4}{c}{TDA} \\ \cline{4-15} 
	                       &                    Mask  & Mask&\noc{80} &\noc{85}&\noc{90}&\iou{1} &\noc{80} &\noc{85} &\noc{90}&\iou{1} &\noc{80} &\noc{85} &\noc{90}&\iou{1}      \\ 
    \midrule
      \multirow{4}{*}{\sbdf}  &\XSolidBrush& \XSolidBrush  &3.15 &4.08 &6.21 &60.14 &6.94 &8.76 &11.76 &35.80 &2.07 &2.47 &3.60 &66.01 \\
         & \Checkmark&\XSolidBrush  &2.45 &3.34 &5.46 &73.99 &6.58 &8.44 &11.51 &42.27 &1.92 &2.27 &3.38 &71.82 \\
        &\XSolidBrush& \Checkmark  &2.49 &3.40 &5.52 &72.87 &6.50 &8.36 &11.43 &42.76 &1.87 &2.23 &3.39 &71.93 \\
        &  \Checkmark& \Checkmark&\textbf{2.41} &\textbf{3.31} &\textbf{5.42} &\textbf{74.74} &\textbf{6.40} &\textbf{8.26} &\textbf{11.36} &\textbf{44.16}  &\textbf{1.84} &\textbf{2.20} &\textbf{3.33} &\textbf{73.12} \\
    \bottomrule
	\end{tabular}
	\caption{
    The ablation study about NoC@X and IoU\&X for introducing different reference masks.
    %
      %
      %
      %
      %
 }
	\label{tab:ablation}
\end{table*}




\myPara{Reference Mask Quality.}
%
We gradually degrade the reference masks into polygons by interval sampling along the boundary points and connecting them.
We adopt these masks as the reference and record the IoU\&1 in \figref{fig:ablation} (a).
We can see that when the mask quality is not too poor, RefCut can always achieve good performance.
This is also in line with expectations, as we use the average feature of the reference mask as the prompt, so the impact of contour segmentation quality on the final performance is relatively small. 
This indicates that in practical use, users only need to provide a rough reference mask to achieve good results.

\myPara{Reference Target Size.}
We keep reducing the size of the reference image and the mask to get the performance of RefCut for the reference target at different scale ratios.
We adopt these changed reference images and masks and record the IoU score with the first click in \figref{fig:ablation} (b).
The results are as expected, the reduction in the size of the reference target leads to a decrease in the performance for RefCut. 
This also matches the results on the PASCAL-Part~\cite{chen2014pascalpart} dataset, which has a large number of very small targets, and using them as reference clearly does not yield good results.

\vspace{-10pt}
\section{Limitation}
\vspace{-5pt}
\label{sec:limit}
For tasks requiring freeform annotation with uncertain categories, general interactive segmentation methods are  more suitable, as our RefCut requires an additional reference image, making it less ideal for such cases. 
RefCut is designed for specific scenarios where users need to annotate a large volume of specific targets. 
By using a reference image, RefCut effectively reduces ambiguity, saving the annotation time for users. 
In addition, when labeling multiple kinds of targets like a horse's head, legs, and tail, RefCut can also handle this scenario. 
In such cases, users provide one reference per label, and when selecting a label before annotation, the model automatically aligns with the relevant reference, accelerating the annotation process for each category.

\vspace{-10pt}
\section{Conclusion}
\vspace{-5pt}
This paper introduces RefCut, an innovative framework for interactive segmentation designed to resolve the common issue of ambiguity when segmenting specific targets. 
It leverages a reference image and corresponding masks provided by the user to guide the model more intuitively, significantly reducing the interactive burden for large-scale annotation with specific objects. 
%
%
Furthermore, we propose the Target Disassembly Dataset, which offers an evaluation benchmark for handling various ambiguous scenarios.
Comprehensive experiments across diverse datasets show that RefCut achieves leading performance, enhancing both accuracy and user control in interactive segmentation.




{
\small
\bibliographystyle{unsrtnat}
\bibliography{reference}
}








\end{document}